\documentclass[10pt,twocolumn,letterpaper]{article}
\usepackage[T1]{fontenc}
\usepackage[final]{cvpr}
\usepackage{tabularx}
\newcommand{\ours}{\textbf{Video-OPSD}\xspace}
\newcommand{\blfootnote}[1]{%
  \begingroup
  \renewcommand{\thefootnote}{}%
  \footnotetext{#1}%
  \endgroup
}
\newcommand{\xww}[1]{}
\newcommand{\hsq}[1]{}
\newcommand{\wzy}[1]{}
\definecolor{revEcolor}{rgb}{0,0,0}

\definecolor{revFcolor}{rgb}{0,0,0}

\definecolor{revGcolor}{rgb}{0,0,0}

\definecolor{revHcolor}{rgb}{0,0,0}

\definecolor{cvprblue}{rgb}{0.21,0.49,0.74}
\usepackage[breaklinks,colorlinks,allcolors=cvprblue]{hyperref}
\def\confName{CVPR}
\def\confYear{2027}
\title{Video-OPSD: Exploiting Privileged Visual Evidence for On-Policy Self-Distillation in Video Large Language Models}
\author{
Ziyue Wang$^{1*}$ \quad Shiqi Huang$^{1*}$ \quad Weiwen Xu$^{2}$ \quad Bihan Wen$^{1}$ \quad Xudong Jiang$^{1\dagger}$\\
$^{1}$School of Electrical and Electronic Engineering, Nanyang Technological University\\
$^{2}$The Chinese University of Hong Kong\\
{\small\texttt{\{ziyue005, shiqi006\}@e.ntu.edu.sg, weiwen.xuu@gmail.com, \{bihan.wen, exdjiang\}@ntu.edu.sg}}
}
\begin{document}
\maketitle
\blfootnote{$^{*}$These authors contributed equally.}
\blfootnote{$^{\dagger}$Corresponding author.}
\begin{abstract}
On-policy self-distillation (OPSD) has recently emerged as an effective post-training paradigm that provides dense token-level supervision from a privileged self-teacher, yet it remains underexplored for Video Large Language Models (Video-LLMs). Existing methods typically construct privileged teachers by augmenting their context with additional information while keeping the primary input unchanged for both teacher and student. In video reasoning, however, privileged information can also be found within the primary input itself: videos contain substantial temporal redundancy, and only a few frames carry the evidence needed to answer a question. Building on this observation, we present \ours, an OPSD framework that exploits privileged visual evidence for both self-teacher construction and knowledge transfer. First, our Evidence-Grounded Self-Teacher conditions the teacher exclusively on annotated evidence frames, while the student reasons over a broader set of frames that contains them. This focused visual context enables more targeted supervision without introducing visual content unavailable to the student. Second, our Evidence-Guided Token Optimization adaptively weights token-level distillation according to an estimate of each reasoning token's reliance on privileged visual evidence, thereby emphasizing perceptually grounded reasoning. Across five video benchmarks and three Qwen-VL backbones, \ours consistently outperforms Standard OPSD by 1.8--2.3 percentage points on average, establishing an effective and efficient post-training approach for Video-LLMs.

\end{abstract}
\section{Introduction}
\label{sec/intro}
On-policy distillation (OPD) has recently emerged as an effective post-training paradigm for providing dense token-level supervision from a teacher model for policy optimization~\cite{agarwal2024onpolicy, gu2024minillm, lu2025onpolicydistillation}. 
Building upon this paradigm, on-policy self-distillation (OPSD) further improves both optimization efficiency and generality by leveraging a self-teacher conditioned on privileged information without requiring an external teacher~\cite{hubotter2026reinforcement, zhao2026self}. While these paradigms have demonstrated promising performance across language models and image tasks, their adaptation to Video Large Language Models (Video-LLMs) remains largely underexplored.

\begin{figure}[t]
  \centering
   \includegraphics[width=\linewidth]{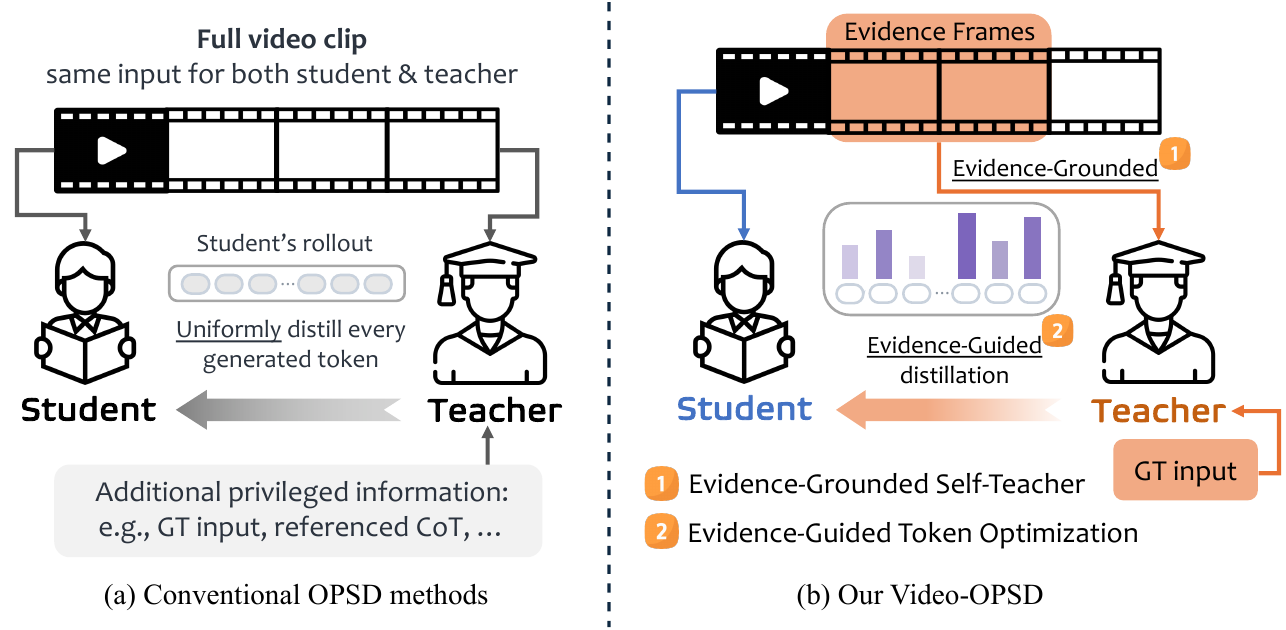}
   \caption{(a) Conventional OPSD, where the teacher and student share the same primary input and token-level distillation is applied uniformly. (b) Our evidence-driven \ours, featuring an Evidence-Grounded Self-Teacher and Evidence-Guided Token Optimization.} 
   \label{fig:teaser}
\end{figure}


Conventional OPSD methods construct a self-teacher by injecting privileged information as additional context, e.g., chain-of-thought references or environmental feedback~\cite{hubotter2026reinforcement, shenfeld2026self, zhao2026self}, while keeping the primary input the same as the student, as shown in Figure \ref{fig:teaser}(a).
The teacher exploits the model's in-context learning capability to make use of the injected privileged information, providing guided supervision during online optimization. While this paradigm has proven effective for textual and image tasks, it overlooks a fundamental property of video understanding. 
Videos inherently contain substantial temporal redundancy~\cite{song2024moviechatdensetokensparse}, where only a small subset of frames provides the visual evidence necessary for answering a question~\cite{xiao2024itrustanswervisually}.
Consequently, both the teacher and student reason over the same video, which contains a mixture of relevant and irrelevant visual information. This shared, unfiltered context limits the teacher's ability to focus on the visual evidence that is actually relevant, thereby degrading the quality of the supervision it provides. This video-specific challenge motivates a different form of privileged supervision in which the teacher can reason from a more focused visual observation.

Based on this insight, we construct an Evidence-Grounded Self-Teacher using asymmetric visual inputs for the teacher and student. The student reasons over a broader set of frames that contains the evidence frames together with additional frames from the rest of the video, whereas the teacher is conditioned only on the \textit{evidence frames} that support the correct answer. By removing temporally redundant and distracting content from the teacher's context, 
the teacher is conditioned on a cleaner, evidence-grounded input rather than having to identify relevant evidence on its own, thereby producing a more accurate and reliable supervisory signal for on-policy optimization.


\begin{figure*}[!t]
    \centering
    \includegraphics[width=0.95\linewidth]{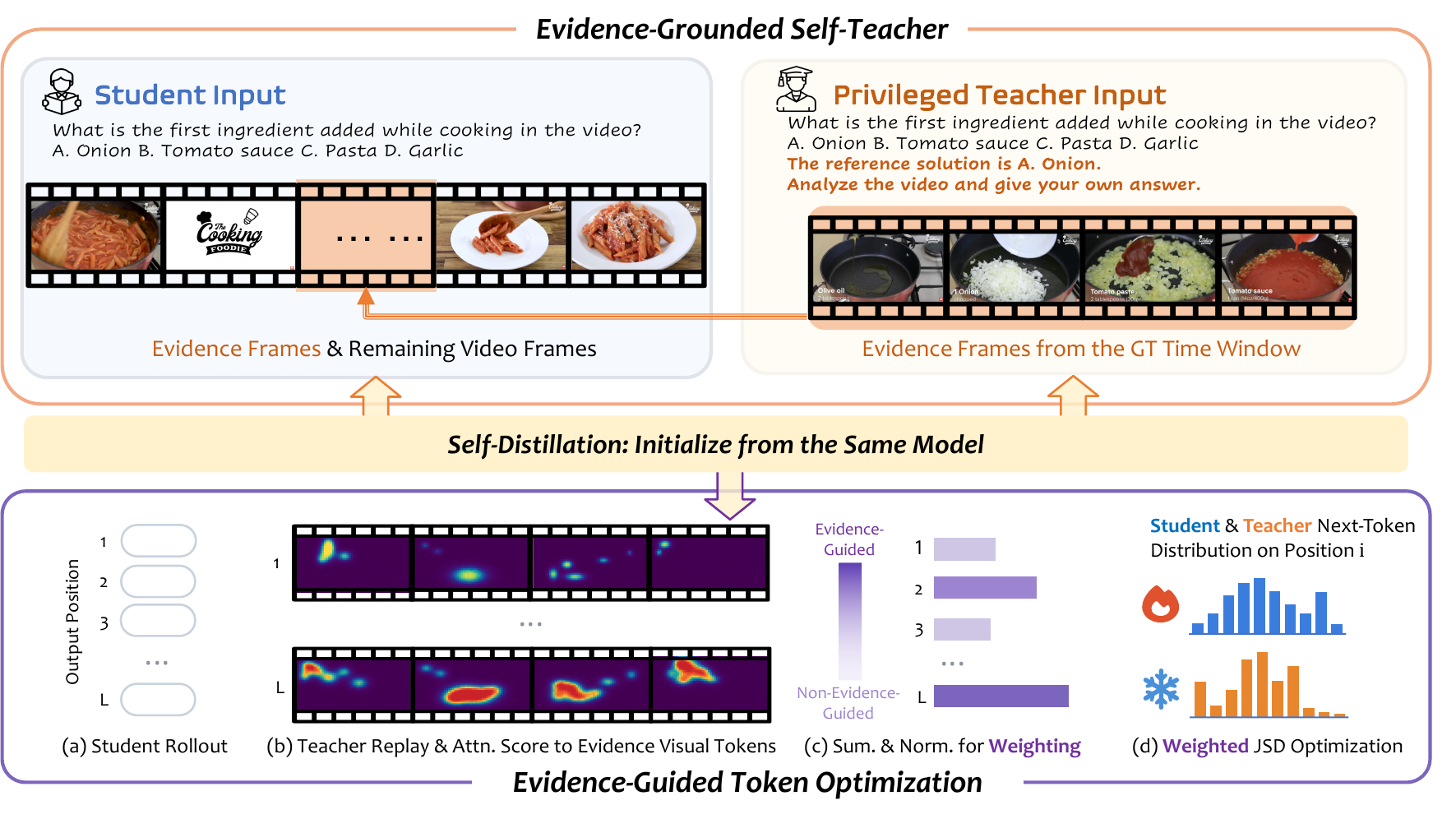}
    \caption{
        Overview of \ours.
        The Evidence-Grounded Self-Teacher observes only the evidence frames together with the gold answer, while the student observes the same evidence frames within a broader video context.
        The student generates a single on-policy trajectory, which is subsequently replayed by the teacher under its privileged context.
        Evidence-Guided Token Optimization measures each generated token's attention to evidence visual signals and uses the resulting scores to weight the token-level JSD objective.
    }
    \label{fig:framework}
\end{figure*}

While an Evidence-Grounded Self-Teacher provides a substantially more reliable
supervisory signal, effectively transferring this privileged knowledge to the student remains an open challenge. 
Existing OPSD uniformly distills the teacher's supervision across all tokens in the student's rollout as in Figure \ref{fig:teaser}(a), implicitly treating each token as equally important throughout the trajectory. To more effectively exploit the teacher's grounded perceptual reasoning capability, the distillation process should reflect how strongly each reasoning token is supported by privileged visual evidence. To this end, we propose Evidence-Guided Token Optimization, which aggregates the teacher's visual attention into token-level visual scores that quantify each token's reliance on privileged visual evidence. These scores are normalized into token-wise weights to modulate the distillation objective, thereby naturally emphasizing perceptually grounded reasoning during on-policy optimization.

Overall, our method rethinks both self-teacher construction and knowledge transfer in OPSD by systematically leveraging privileged visual evidence, resulting in our evidence-driven Video-OPSD illustrated in Figure~\ref{fig:teaser}(b). Through the proposed Evidence-Grounded Self-Teacher and Evidence-Guided Token Optimization, our framework constructs a more informative teacher from task-relevant frames and selectively transfers its perceptual reasoning capability to the student. Extensive experiments on video understanding and reasoning benchmarks demonstrate that our approach consistently outperforms the OPSD baseline.

Our contributions are summarized as follows:
\begin{itemize}
    \item We propose \ours, an on-policy self-distillation framework for Video-LLMs that systematically exploits privileged visual evidence through both evidence-grounded teacher construction and evidence-guided knowledge transfer.
    \item We introduce \textbf{Evidence-Grounded Self-Teacher}, which conditions the teacher only on evidence frames, enabling more reliable 
    supervision from focused visual context.
    \item We introduce \textbf{Evidence-Guided Token Optimization}, which adaptively weights token-level distillation according to each token's reliance on privileged visual evidence, enhancing perceptually grounded reasoning.
    \item Extensive experiments across multiple video understanding and reasoning benchmarks validate the effectiveness of our framework, which consistently outperforms Standard OPSD and closes its gap to GRPO. Moreover, \ours remains robust to the quality of evidence annotations, maintaining comparable accuracy under both perturbed evidence windows and windows predicted by the backbone itself, which suggests its applicability to videos without human temporal annotations.
\end{itemize}

\section{Related Work}
\label{sec:related_work}

\subsection{Reinforcement Learning for Video-LLMs}
Reinforcement learning has recently emerged as an effective paradigm for enhancing the reasoning capabilities of Video-LLMs. Building on reinforcement learning with verifiable rewards (RLVR) ~\cite{shao2024deepseekmath, deepseekai2025deepseekr1incentivizingreasoningcapability}, recent methods introduce video-specific optimization strategies, including temporal sensitivity and consistency~\cite{feng2025video,huang2026evovid}, spatio-temporal reasoning~\cite{li2025videochat,wang2025time,wang2026video,meng2026openo3videogroundedvideoreasoning}, and detailed perception reasoning~\cite{du2026appoattentionguidedperceptionpolicy}, to better align policy learning with video understanding. More recently, a few studies have explored on-policy distillation for temporal video grounding~\cite{li2026video} and video reasoning~\cite{lin2026visd}, complementing sparse reward supervision with dense token-level learning signals. Despite this progress, existing methods have yet to exploit the privileged visual evidence available during training to strengthen token-level supervision for perceptually grounded reasoning.

\subsection{On-Policy Distillation and Self-Distillation}
On-policy distillation (OPD) has recently been widely explored for providing dense token-level supervision from a teacher model for policy optimization~\cite{agarwal2024onpolicy, gu2024minillm, lu2025onpolicydistillation, xu2026beyond,  li2026rethinkingonpolicydistillationlarge, li2026video}. Building upon this paradigm, recent on-policy self-distillation (OPSD) methods eliminate the need for an external teacher by constructing a privileged self-teacher from the same policy~\cite{hubotter2026reinforcement, zhao2026self}. 
Existing approaches typically realize privileged supervision by enriching the teacher with additional information ~\cite{zhang2026opsdlonpolicyselfdistillationlongcontext}, such as reference reasoning traces or environmental feedback~\cite{xu2024reasons,yang2026self, shenfeld2026self, lin2026visd,xu2026beyond}. Our work explores a complementary direction for Video-LLMs. Rather than augmenting the teacher with additional context, we exploit annotated visual evidence within temporally redundant videos ~\cite{wang2025retakereducingtemporalknowledge, song2024moviechatdensetokensparse, zou2024secondshoursreviewingmultimodal} to develop an evidence-driven OPSD framework for Video-LLMs that more effectively transfers the teacher's privileged visual knowledge.
\section{Method}
\label{sec/method}

\subsection{Preliminaries: On-Policy Self-Distillation}

In OPSD, a trainable student $\pi_{\theta_S}$ and a frozen self-teacher $\pi_{\theta_T}$ are initialized from the same pretrained Video-LLM. The self-teacher derives its supervision from privileged context available only during training. Specifically, the student is conditioned on $c^{S}$, while the self-teacher is conditioned on the privilege-enhanced context
$c^{T}$.

For each example selected for distillation, the student generates one on-policy trajectory:
\begin{equation}
\hat{y}
=
(\hat{y}_1,\ldots,\hat{y}_L)
\sim
\pi_{\theta_S}(\cdot\mid c^{S}).
\label{eq:student_rollout}
\end{equation}
The self-teacher then replays the student trajectory under its privileged context. At output position $l$, both branches evaluate the same student-generated prefix:
\begin{align}
p_l^{S}
&=
\pi_{\theta_S}
\left(
\cdot\mid c^{S},\hat{y}_{<l}
\right),
\label{eq:student_distribution}
\\
p_l^{T}
&=
\pi_{\theta_T}
\left(
\cdot\mid c^{T},\hat{y}_{<l}
\right).
\label{eq:teacher_distribution}
\end{align}
This prefix alignment provides supervision at states visited by the current student policy. Through teacher-forced replay of the student trajectory, the self-teacher produces next-token distributions for all positions in a single forward, without autoregressively generating a response. Unlike GRPO~\cite{shao2024deepseekmath}, which typically requires multiple rollouts per input, OPSD obtains dense token-level supervision from one student rollout followed by one teacher replay.

\subsection{Framework Overview}

We consider a video reasoning training example:
\begin{equation}
x=(V,q,a,\mathcal{T}^{e}),
\label{eq:training_example}
\end{equation}
where $V$ denotes the video, $q$ is the question, $a$ is the gold answer, and $\mathcal{T}^{e}$ is the annotated temporal interval containing the visual evidence that supports the answer. We initialize a trainable student $\pi_{\theta_S}$ and a frozen self-teacher $\pi_{\theta_T}$ from the same pretrained Video-LLM.

As illustrated in Figure~\ref{fig:framework}, \ours contains two complementary components. First, the Evidence-Grounded Self-Teacher constructs privileged supervision by restricting the teacher's visual observation to frames from the annotated evidence interval. The same evidence frames are included unchanged in the student input, ensuring that the teacher does not rely on visual information unavailable to the student. In addition to the evidence frames, the student observes frames sampled from the remaining video, which encourages it to identify and exploit the evidence within this broader temporal context.

Second, Evidence-Guided Token Optimization uses the teacher's attention during replay to estimate how strongly each generated token relies on visual evidence signals. Since the teacher's entire visual input consists of evidence frames, its attention to visual tokens reflects the attention allocated to visual evidence signals and provides a model-internal proxy for token-level visual evidence reliance. The resulting scores are used to adaptively weight the matching of the token-level distribution.

\subsection{Evidence-Grounded Self-Teacher}

Let $\mathcal{I}^{e}$ denote the indices of frames sampled from the
annotated evidence interval $\mathcal{T}^{e}$, and let
$\mathcal{I}^{a}$ denote the indices of additional frames sampled from
the remaining video. We define
$\mathcal{I}^{S}=\mathcal{I}^{e}\cup\mathcal{I}^{a}$ and construct the
corresponding visual inputs as:
\begin{equation}
V^{T}=V[\mathcal{I}^{e}],
\qquad
V^{S}=V[\mathcal{I}^{S}].
\label{eq:visual_inputs}
\end{equation}
The teacher observes only the evidence frames, whereas the student observes the same evidence frames together with additional frames. Consequently, every visual signal available to the teacher is also accessible to the student, preventing information leakage \cite{yang2026self}. The teacher's visual privilege arises from removing non-evidence context, rather than from observing signals unavailable to the student.

Following prior OPSD methods, the teacher is additionally conditioned on the gold answer:
\begin{equation}
c^{S}=(V^{S},q),
\qquad
c^{T}=(V^{T},q,a).
\label{eq:student_teacher_contexts}
\end{equation}
The gold answer provides textual privilege, while the evidence-grounded observation provides video-specific visual privilege. Together, they enable the teacher to produce supervision from a more focused perceptual context.

Since privileged conditioning does not guarantee a reliable teacher prediction, we follow common practice and apply a correctness gate before distillation. The frozen teacher first greedily generates:
\begin{equation}
\widetilde{y}^{T}
=
\operatorname{Greedy}
\left[
\pi_{\theta_T}(\cdot\mid c^{T})
\right],
\label{eq:teacher_greedy}
\end{equation}
and the example is retained only if:
\begin{equation}
g(x)
=
\mathbb{I}
\left[
\operatorname{ANS}(\widetilde{y}^{T})=a
\right],
\label{eq:teacher_gate}
\end{equation}
where $\operatorname{ANS}$ is a rule-based answer extractor. Incorrect or unparsable teacher responses are excluded from distillation.

\subsection{Evidence-Guided Token Optimization}
To more effectively leverage the teacher's grounded perceptual reasoning, we estimate each generated token's reliance on visual evidence from the teacher's attention during trajectory replay and use the resulting scores to weight the distillation objective. Let $\mathcal{V}^{e}$ denote the indices of visual tokens corresponding to $V^{T}$, let $R$ denote the number of Transformer layers, and let $H$ be the number of attention heads. The evidence score at position $l$ is defined as:
\begin{equation}
s_l
=
\frac{1}{3H}
\sum_{r=R-2}^{R}
\sum_{h=1}^{H}
\sum_{j\in\mathcal{V}^{e}}
A_{l}^{(r,h,j)},
\label{eq:evidence_attention}
\end{equation}
where $A_{l}^{(r,h,j)}$ denotes the teacher's attention from output position $l$ to visual token $j$ at layer $r$ and head $h$. We sum the attention mass over the evidence visual tokens $\mathcal{V}^{e}$ and average it across all heads of the last three layers.

We convert the evidence scores into normalized token weights as below:
\begin{equation}
w_l
=
L
\frac{
s_l
}{
\sum_{u=1}^{L} s_u
},
\label{eq:token_weight}
\end{equation}
\ie, each token receives a share of the distillation weight proportional to its evidence score. Tokens with stronger evidence reliance receive greater distillation weight, while all generated tokens remain part of the objective.

Since an LLM's probability mass is typically concentrated within a small number of top tokens, the Top-K truncation introduces only a small approximation error in practice while substantially reducing computation. Therefore, we compute the token-level divergence over the teacher's Top-$K$ vocabulary support at each token position $l$:
\begin{equation}
\mathcal{K}_l^{T}
=
\operatorname{TopK}_{K}(p_l^{T}),
\label{eq:teacher_topk}
\end{equation}
and renormalize the student and teacher distributions as:
\begin{equation}
\bar{p}_{l,k}^{b}
=
\frac{
p_{l,k}^{b}
}{
\sum_{v\in\mathcal{K}_l^{T}}p_{l,v}^{b}
},
\qquad
k\in\mathcal{K}_l^{T},
\quad
b\in\{S,T\}.
\label{eq:topk_normalization}
\end{equation}
Following the generalized JSD formulation in~\citet{agarwal2024onpolicy}, we define:
\begin{equation}
m_l
=
\beta\bar{p}_l^{T}
+
(1-\beta)\bar{p}_l^{S},
\label{eq:jsd_mixture}
\end{equation}
and
\begin{align}
d_l^{\mathrm{JSD}\text{-}K}
={}&
\beta
D_{\mathrm{KL}}
\left(
\bar{p}_l^{T}\Vert m_l
\right)
\nonumber\\
&+
(1-\beta)
D_{\mathrm{KL}}
\left(
\bar{p}_l^{S}\Vert m_l
\right).
\label{eq:jsd_topk}
\end{align}

The final objective is:
\begin{equation}
\mathcal{L}_{\mathrm{Video\text{-}OPSD}}
=
\mathbb{E}_{x\sim\mathcal{D}}
\left[
g(x)
\cdot
\frac{1}{L}
\sum_{l=1}^{L}
w_l
d_l^{\mathrm{JSD}\text{-}K}
\right].
\label{eq:overall_objective}
\end{equation}
The teacher distributions and evidence-guided weights are treated as fixed targets, and gradients are propagated only through the student branch.


\section{Experiments}
\label{sec/experiments}

\subsection{Experimental Setup}

\paragraph{Training data.}
We construct a compact training set from two datasets with ground-truth temporal window annotations. Specifically, we use 5K randomly sampled examples from STAR~\cite{wu2024starbenchmarksituatedreasoning} and approximately 1.5K examples from the Video-Holmes training split~\cite{cheng2025videoholmesmllmthinklike}. We use the original questions and gold answers provided by the two datasets and do not introduce additional human annotations.

\paragraph{Evaluation benchmarks.}
We evaluate all methods on five video understanding and reasoning benchmarks: Video-Holmes~\cite{cheng2025videoholmesmllmthinklike}, Video-MMMU~\cite{hu2025videommmuevaluatingknowledgeacquisition}, Video-MME~\cite{fu2025videommefirstevercomprehensiveevaluation}, TempCompass~\cite{liu2024tempcompassvideollmsreally}, and WorldSense~\cite{hong2026worldsenseevaluatingrealworldomnimodal}. All benchmarks are evaluated under a unified video-only protocol with 32 frames: the model receives only video frames and the textual question, without audio, subtitles, or transcripts. For Qwen2.5-VL-7B-Instruct, we follow the evaluation prompt of Video-R1~\cite{feng2025video}, which requests reasoning within \texttt{<think>} tags. Since Qwen3-VL models frequently produce malformed outputs under this format, we evaluate them with a plain-text reasoning prompt that ends with an \texttt{<answer>} tag. Within each backbone, all methods are evaluated with the same prompt.

\begin{table*}[!t]
    \centering
    \caption{Main results on five video understanding and reasoning benchmarks across three Qwen-VL backbones under the unified video-only evaluation protocol. Comparisons are made within each backbone. Results are averaged over three independent runs.}
    \label{tab:main_results}
    \small
    \setlength{\tabcolsep}{7.0pt}
    \begin{tabular}{lcccccc}
        \toprule
        Method
        & Video-Holmes
        & Video-MMMU
        & Video-MME
        & TempCompass
        & WorldSense
        & Avg. \\
        \midrule

        Qwen2.5-VL-7B-Instruct
        & 27.8
        & 48.1
        & 57.6
        & 67.9
        & 36.1
        & 47.5 \\

        \quad + SFT
        & 33.2
        & 48.5
        & 56.4
        & 69.2
        & 36.3
        & 48.7 \\

        \quad + GRPO
        & 35.3
        & 49.9
        & \textbf{60.7}
        & 70.9
        & 36.6
        & 50.7 \\

        \quad + OPSD
        & 33.4
        & 48.1
        & 57.7
        & 68.5
        & 36.7
        & 48.9 \\

        \quad + \ours
        & \textbf{36.2}
        & \textbf{49.9}
        & 58.6
        & \textbf{70.9}
        & \textbf{39.2}
        & \textbf{51.0} \\

        \midrule

        Qwen3-VL-8B-Instruct
        & 41.0
        & 63.6
        & 63.5
        & 74.4
        & 40.6
        & 56.6 \\

        \quad + SFT
        & 41.3
        & 65.8
        & 62.7
        & 75.5
        & 40.8
        & 57.2 \\

        \quad + GRPO
        & 45.9
        & \textbf{68.7}
        & \textbf{66.9}
        & 77.2
        & \textbf{42.3}
        & \textbf{60.2} \\

        \quad + OPSD
        & 42.6
        & 67.3
        & 65.5
        & 75.1
        & 40.5
        & 58.2 \\

        \quad + \ours
        & \textbf{46.4}
        & 68.1
        & 66.1
        & \textbf{77.5}
        & 42.0
        & 60.0 \\

        \midrule

        Qwen3-VL-4B-Instruct
        & 36.5
        & 58.2
        & 59.6
        & 71.7
        & 38.4
        & 52.9 \\

        \quad + SFT
        & 39.3
        & 55.8
        & 57.4
        & 72.2
        & 39.0
        & 52.7 \\

        \quad + GRPO
        & \textbf{44.2}
        & 60.2
        & \textbf{60.3}
        & 72.6
        & \textbf{40.8}
        & 55.6 \\

        \quad + OPSD
        & 40.2
        & 57.5
        & 58.4
        & 71.9
        & 39.0
        & 53.4 \\

        \quad + \ours
        & 43.8
        & \textbf{61.0}
        & 60.1
        & \textbf{72.9}
        & 40.5
        & \textbf{55.7} \\

        \bottomrule
    \end{tabular}
\end{table*}

\paragraph{Implementation details.}
We conduct the main experiments on three models: Qwen2.5-VL-7B-Instruct~\cite{bai2025qwen25vltechnicalreport}, Qwen3-VL-4B-Instruct~\cite{bai2025qwen3vltechnicalreport}, and Qwen3-VL-8B-Instruct~\cite{bai2025qwen3vltechnicalreport}. Training is conducted for one epoch on a total of 6.5K training examples. The teacher receives 8 frames sampled from the annotated evidence window, while the student receives the same 8 frames together with 8 frames sampled from the remaining video. We adopt JSD-Top-50 as the distillation objective with $\beta=0.5$. The per-device batch size is set to 2 with 8 gradient accumulation steps, and the learning rate is $1\times10^{-6}$. Training is performed in BF16 precision with gradient checkpointing and DeepSpeed ZeRO-2 enabled. GRPO uses a group size of $G=8$ and a KL coefficient of $0.01$.

\paragraph{Baselines.}
For each backbone, we compare our method against supervised fine-tuning (SFT), GRPO~\cite{shao2024deepseekmath}, and Standard OPSD~\cite{zhao2026self}. All the methods are conducted with the same training data, frame budget, and evaluation protocol. For the SFT baseline, we condition the backbone on the ground-truth answer and use its greedily decoded reasoning trajectory as the supervision target for training.
Standard OPSD conditions the frozen teacher on the gold answer, provides it with the same visual input as the student, uniformly weights all token-level distillation losses, and applies the same teacher correctness gate as \ours.

\subsection{Main Results}

Table~\ref{tab:main_results} reports the results across three Qwen-VL backbones. We report the average results over three independent runs. \ours consistently improves the base models, yielding average gains of $3.5$, $3.4$, and $2.8$ points on Qwen2.5-VL-7B-Instruct, Qwen3-VL-8B-Instruct, and Qwen3-VL-4B-Instruct, respectively. More importantly, compared to Standard OPSD, our method further improves the average score by $2.1$, $1.8$, and $2.3$ points and achieves higher performance in all $15$ backbone--benchmark comparisons. These consistent gains demonstrate that exploiting privileged visual evidence provides effective supervision beyond vanilla OPSD.

\begin{figure*}[t]
  \centering
  \includegraphics[width=\linewidth]{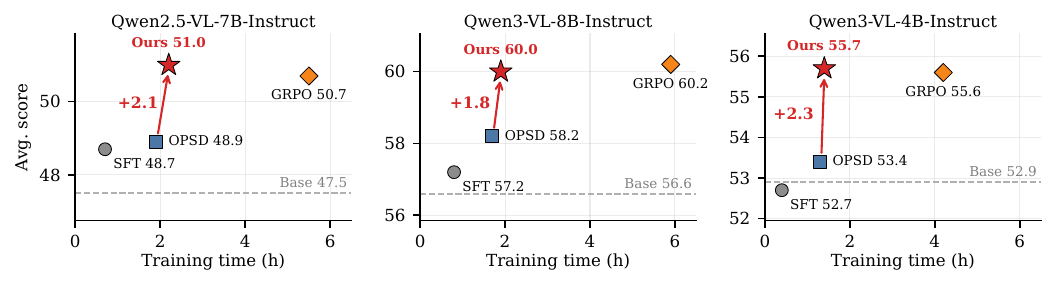}
  \caption{Average score versus training time on three backbones. Training time covers the full run, including the teacher correctness gate.}
  \label{fig:acc_time}
\end{figure*}

Moreover, whereas Standard OPSD trails GRPO by $1.8$, $2.0$, and $2.2$ average points on the three backbones, \ours closes this gap with average scores within $0.3$ points of GRPO ($51.0$ \vs $50.7$, $60.0$ \vs $60.2$, and $55.7$ \vs $55.6$). As shown in Figure~\ref{fig:acc_time}, this is achieved at a training cost comparable to that of Standard OPSD ($2.2$ \vs $1.9$, $1.9$ \vs $1.7$, and $1.4$ \vs $1.3$ hours), requiring at most $40\%$ of the training time of GRPO. These results indicate that privileged visual evidence allows \ours to match the performance of GRPO at a training cost comparable to that of OPSD.

\subsection{Ablation Studies}
\label{sec:ablation}

All ablation experiments in this section are conducted using Qwen2.5-VL-7B-Instruct. Unless otherwise specified, all variants use the same training data, frame budget, frozen teacher, correctness gate, Top-50 JSD objective, and evaluation protocol. This controlled setup ensures that performance differences can be primarily attributed to the design factor under investigation in each ablation. We report average performance across the five evaluation benchmarks to provide a concise and consistent comparison among variants. We first analyze the contribution of each component (RQ1), then examine the design of the Evidence-Grounded Self-Teacher (RQ2--RQ4) and Evidence-Guided Token Optimization (RQ5--RQ6), and finally assess the robustness of \ours to the quality of evidence annotations (RQ7).

\paragraph{RQ1: How does each component contribute to the performance of \ours?}

We first conduct a controlled $2\times2$ component ablation to disentangle the contributions of the Evidence-Grounded Self-Teacher and Evidence-Guided Token Optimization. As shown in Table~\ref{tab:component_ablation}, the Evidence-Grounded Self-Teacher alone improves the average score of Standard OPSD from $48.9$ to $50.0$, while token weighting provides a larger improvement to $50.4$. Combining both components further increases the average score to $51.0$, outperforming Standard OPSD by $2.1$ points.

These results indicate that the two designs play distinct but complementary roles: the Evidence-Grounded Self-Teacher improves the quality of privileged supervision, while Evidence-Guided Token Optimization transfers this supervision more effectively by emphasizing tokens that rely on visual evidence.

\begin{table}[t]
    \centering
    \caption{Component ablation of the Evidence-Grounded Self-Teacher and Evidence-Guided Token Optimization (average over five benchmarks).}
    \label{tab:component_ablation}
    \small
    \setlength{\tabcolsep}{18.0pt}
    \begin{tabular}{lc}
        \toprule
        Method & Avg. \\
        \midrule
        Qwen2.5-VL-7B-Instruct & 47.5 \\
        \midrule
        OPSD & 48.9 \\
        \quad + Evidence-Grounded Self-Teacher & 50.0 \\
        \quad + Evidence-Guided Token Optimization & 50.4 \\
        \ours (Both) & \textbf{51.0} \\
        \bottomrule
    \end{tabular}
\end{table}

\paragraph{RQ2: Where does the privileged teacher's advantage come from?}

The privileged teacher differs from the student in two aspects: it observes temporally localized evidence frames and is conditioned on the gold answer. To disentangle these two sources of privilege, we conduct a $3 \times 2$ ablation over the teacher input. For the visual input, we compare three configurations: (1) eight frames sampled from the annotated temporal evidence window, (2) the same eight evidence frames with additional eight non-evidence context frames to form a 16-frame input, and (3) 16 uniformly sampled frames from the full video. Together, configuration (2) serves as a bridge: it isolates the effect of added context frames relative to configuration (1), and controls for frame count relative to configuration (3), which isolates the effect of temporal localization. For each visual configuration, we further evaluate the teacher both with and without gold-answer conditioning.

As shown in Table~\ref{tab:privileged_source_ablation}, temporally focused visual evidence consistently produces a stronger teacher. Without gold-answer conditioning, using only evidence frames achieves $70.4\%$ inference accuracy. Adding non-evidence context frames reduces the accuracy to $64.9\%$, while using 16 uniformly sampled frames further decreases it to $49.8\%$. Therefore, compared with uniform sampling, temporally localized evidence frames improve teacher accuracy by $20.6$ percentage points. The same ordering remains after introducing gold-answer conditioning. Although gold-answer conditioning improves teacher accuracy under every visual input configuration, it does not eliminate the degradation caused by non-evidence visual content. These results show that the privileged teacher's advantage comes from both answer-level conditioning and evidence-grounded visual input, with temporal evidence localization providing additional benefits beyond simply increasing the number of observed frames.

\begin{table}[t]
    \centering
    \caption{
    Accuracy on the training set to disentangle the sources of privileged teacher information.
    }
    \label{tab:privileged_source_ablation}
    \small
    \setlength{\tabcolsep}{1.0pt}
    \begin{tabularx}{\columnwidth}{>{\raggedright\arraybackslash}Xcc}
        \toprule
        Teacher Visual Input
        & \shortstack{Gold\\Answer}
        & \shortstack{Accuracy\\(\%)} \\
        \midrule

        Evidence Frames Only
        & No
        & \textbf{70.4} \\

        Evidence Frames + 8 Non-Evidence Frames
        & No
        & 64.9 \\

        Uniformly Sampled 16 Frames
        & No
        & 49.8 \\

        \midrule

        Evidence Frames Only
        & Yes
        & \textbf{85.3} \\

        Evidence Frames + 8 Non-Evidence Frames
        & Yes
        & 83.7 \\

        Uniformly Sampled 16 Frames
        & Yes
        & 78.5 \\

        \bottomrule
    \end{tabularx}
\end{table}

\paragraph{RQ3: How do non-evidence frames affect privileged supervision?}
\label{sec:non_evidence_frame_ablation}
We augment the teacher's evidence frames with four or eight frames sampled outside the annotated window. As shown in Table~\ref{tab:non_evidence_frames}, the average score drops from $51.0$ to $50.6$ and $50.4$, respectively. Together with RQ2, where temporal localization yields a stronger teacher even at a matched frame count, this indicates that the benefit of the Evidence-Grounded Self-Teacher stems from removing temporally irrelevant content.

\begin{table}[t]
    \centering
    \caption{Effect of adding non-evidence frames to the teacher input (average over five benchmarks).}
    \label{tab:non_evidence_frames}
    \small
    \setlength{\tabcolsep}{10.0pt}
    \begin{tabular}{lc}
        \toprule
        Non-evidence frames added to the teacher & Avg. \\
        \midrule
        $0$ (evidence frames only, ours) & \textbf{51.0} \\
        $4$ & 50.6 \\
        $8$ & 50.4 \\
        \bottomrule
    \end{tabular}
\end{table}

\paragraph{RQ4: Does updating the self-teacher improve distillation?}
\label{sec:teacher_update_ablation}
We compare the frozen teacher with teachers updated from the student via an exponential moving average (EMA) with rates $0.01$ and $0.03$. As shown in Table~\ref{tab:teacher_update}, the frozen teacher performs best ($51.0$ \vs $50.2$ and $50.3$). We attribute this to the non-stationary targets introduced by updating the teacher toward the student, whereas a frozen teacher provides stable supervision throughout training.

\begin{table}[t]
    \centering
    \caption{Effect of updating the self-teacher (average over five benchmarks).}
    \label{tab:teacher_update}
    \small
    \setlength{\tabcolsep}{12.0pt}
    \begin{tabular}{lc}
        \toprule
        Teacher update & Avg. \\
        \midrule
        EMA ($0.01$) & 50.2 \\
        EMA ($0.03$) & 50.3 \\
        Frozen (ours) & \textbf{51.0} \\
        \bottomrule
    \end{tabular}
\end{table}

\paragraph{RQ5: How does Evidence-Guided Token Optimization compare
with alternative token weighting strategies?}

We compare several strategies for allocating token-level distillation signals. Uniform weighting assigns equal weight to every generated token position.
Motivated by TIP~\cite{xu2026tiptokenimportanceonpolicy},
Token Entropy uses the student's predictive entropy as a
generic measure of token importance. The entropy scores are then converted into soft token weights. We additionally evaluate a hard evidence-selection strategy,
Evidence-Guided (Top-$30\%$), which updates only the $30\%$ of token positions with the highest evidence scores $s_l$. Evidence-Guided (Soft), \ie, our Evidence-Guided Token Optimization, retains all generated positions and weights their distillation losses in proportion to their evidence scores (Eq.~\ref{eq:token_weight}).

As reported in Table~\ref{tab:token_weighting}, token-entropy weighting improves the average score from $50.0$ to $50.3$, confirming that student uncertainty provides a useful optimization signal. However, the signal is generic and does not explicitly measure whether a generated token relies on the privileged visual evidence. Hard top-$30\%$ selection obtains an average score of $50.1$. In comparison, soft evidence-guided weighting achieves the highest average score of $51.0$.
It outperforms uniform weighting, token-entropy weighting, and hard top-$30\%$ selection by $1.0$, $0.7$, and $0.9$ average points, respectively. These results support softly allocating distillation strength according to the relevance of privileged visual evidence rather than relying solely on generic uncertainty or hard token filtering.

\begin{table}[t]
    \centering
    \caption{Comparison of token weighting strategies (average over five benchmarks).}
    \label{tab:token_weighting}
    \small
    \setlength{\tabcolsep}{10.0pt}
    \begin{tabular}{lc}
        \toprule
        Weighting strategy & Avg. \\
        \midrule
        Uniform & 50.0 \\
        Token Entropy & 50.3 \\
        Evidence-Guided (Top-$30\%$) & 50.1 \\
        Evidence-Guided (Soft, ours) & \textbf{51.0} \\
        \bottomrule
    \end{tabular}
\end{table}

\begin{figure}[t]
    \centering
    \begin{subfigure}{\columnwidth}
        \centering
        \includegraphics[width=\linewidth]{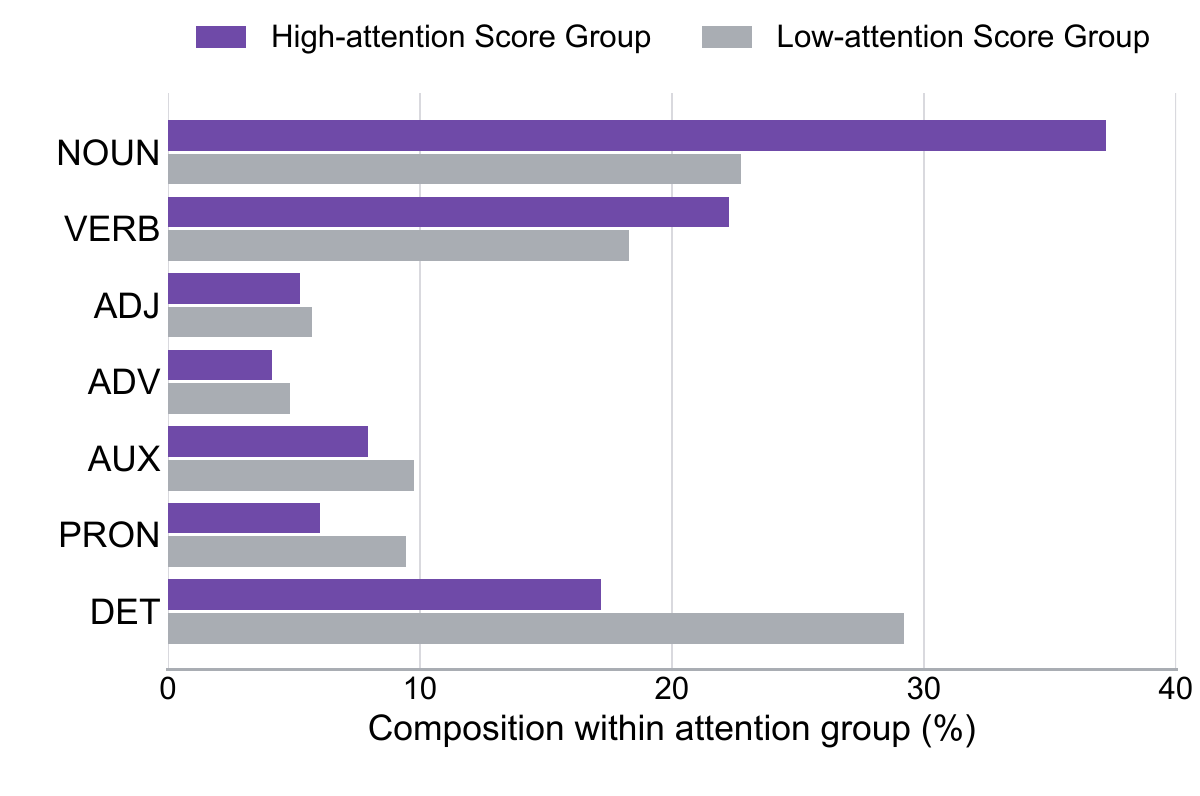}
        \caption{Part-of-speech composition analysis of high- and low-attention tokens.}
        \label{fig:pos_attention_pattern_a}
    \end{subfigure}
    \vspace{2pt}
    \begin{subfigure}{\columnwidth}
        \centering
        \includegraphics[width=0.9\linewidth]{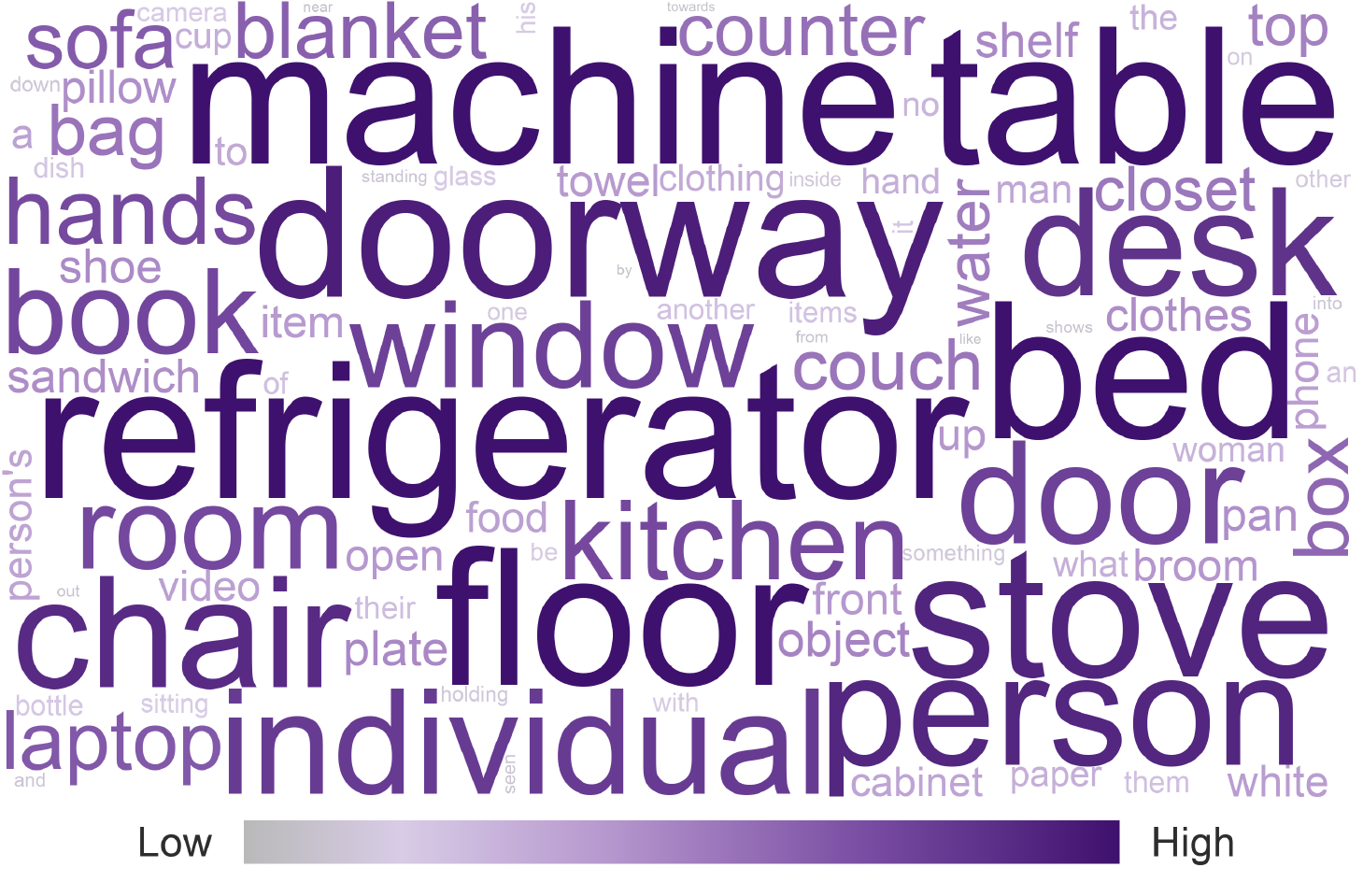}
        \caption{Word cloud of generated tokens, with size and color indicating their mean distillation weights.}
        \label{fig:pos_attention_pattern_b}
    \end{subfigure}
    \caption{Analysis of the tokens emphasized by Evidence-Guided Token Optimization.}
    \label{fig:pos_attention_pattern}
\end{figure}

\paragraph{RQ6: Which tokens does evidence-guided weighting emphasize?}
\label{sec:pos_attention_analysis}
For each student trajectory, we rank word-level tokens by their evidence score $s_l$ and assign the top $50\%$ to a high-attention group and the bottom $50\%$ to a low-attention group. We consider seven part-of-speech categories (nouns, verbs, adjectives, adverbs, auxiliary verbs, pronouns, and determiners), excluding punctuation and other categories, and aggregate the tokens across trajectories. As shown in Figure~\ref{fig:pos_attention_pattern}(a), the high-attention group contains more nouns and verbs, whereas determiners, pronouns, and auxiliary verbs dominate the low-attention group. To further inspect individual tokens, we group identical generated tokens and visualize their mean normalized distillation weights $w_l$ in Figure~\ref{fig:pos_attention_pattern}(b). Tokens referring to visually observable entities and scene elements, such as \emph{table}, \emph{refrigerator}, \emph{doorway}, \emph{desk}, \emph{bed}, and \emph{stove}, receive relatively high weights, whereas generic function words and weakly grounded expressions receive smaller weights. Evidence-guided weighting thus emphasizes content words describing visual entities and actions, consistent with its goal of strengthening perceptually grounded reasoning.

\paragraph{RQ7: How robust is \ours to the quality of evidence annotations?}
\label{sec:evidence_localization_robustness}
We replace the annotated evidence window with three types of alternatives: perturbed annotations, expanded by $20\%$ on both sides or shifted by $20\%$ of their duration; windows predicted by GPT-5~\cite{openai2026gpt5} or the base Qwen2.5-VL-7B-Instruct model from each training video sampled at 1 fps (at most 256 frames); and random windows of the same length as the annotated ones. In each variant, both the teacher's frames and the student's evidence frames are sampled from the replacement window.

As shown in Table~\ref{tab:window_perturbation}, perturbed windows reduce accuracy only slightly, from $51.0$ to $50.7$ and $50.5$. Predicted windows achieve comparable accuracy ($50.8$ and $50.6$) despite their low IoU ($0.53$ and $0.42$): IoU also penalizes windows that extend beyond the annotation, whereas these windows still cover most of the annotated evidence ($73\%$ and $67\%$). In contrast, random windows cover only $13\%$ of the evidence and reduce accuracy to $47.9$. These results indicate that the effectiveness of \ours depends on whether the window covers the evidence rather than on its precise boundaries, and that \ours remains effective when the windows are predicted by the backbone itself.

\begin{table}[t]
    \centering
    \caption{Robustness of \ours to evidence-window quality (average over five benchmarks). IoU and coverage (Cov.) are computed against the ground-truth window. $^\dagger$Nominal values.}
    \label{tab:window_perturbation}
    \small
    \setlength{\tabcolsep}{4.0pt}
    \begin{tabular}{lccc}
        \toprule
        Evidence window & IoU & Cov. & Avg. \\
        \midrule
        Ground-truth window & 1.00 & 100\% & \textbf{51.0} \\
        Window expanded by $20\%$ & 0.71$^\dagger$ & 100\%$^\dagger$ & 50.7 \\
        Window shifted by $20\%$ & 0.67$^\dagger$ & 80\%$^\dagger$ & 50.5 \\
        \midrule
        Predicted by GPT-5 & 0.53 & 73\% & 50.8 \\
        Predicted by Qwen2.5-VL-7B-Instruct & 0.42 & 67\% & 50.6 \\
        \midrule
        Random window & 0.10 & 13\% & 47.9 \\
        \bottomrule
    \end{tabular}
\end{table}

\section{Conclusion}

We introduce \ours, an evidence-driven on-policy self-distillation framework that enhances video reasoning by exploiting privileged visual evidence during training. Our framework constructs an Evidence-Grounded Self-Teacher from temporally localized evidence frames and introduces Evidence-Guided Token Optimization to selectively transfer supervision according to each reasoning token's reliance on visual evidence. Experiments across five video understanding and reasoning benchmarks demonstrate consistent improvements over supervised fine-tuning and Standard OPSD, and show that \ours closes the gap between Standard OPSD and GRPO while keeping the training cost close to that of Standard OPSD. Extensive ablations further validate that evidence-grounded teacher construction improves the quality of privileged supervision and that evidence-guided weighting enables more effective knowledge transfer. Moreover, \ours remains effective when evidence windows are predicted by the backbone itself. One limitation of \ours is its reliance on temporal evidence windows during training: even when predicted by the backbone itself rather than annotated by humans, these windows require an additional one-time inference cost over the training videos. Overall, our results highlight privileged visual evidence as a valuable source of supervision for efficient post-training of Video-LLMs.

\section*{LLM Usage Statement}
\label{sec:llm_usage}
We used LLM-based assistants in the following ways.
\textbf{Code.} LLM coding assistants (\textit{Codex}) were used to help write and debug parts of the training and evaluation code. All generated code was reviewed and tested by the authors, and all reported results come from experiments run by the authors.
\textbf{Writing and figures.} An LLM assistant (Chat-GPT) was used to migrate the manuscript to the CVPR template, polish the language, suggest revisions to the presentation and organization of the paper.
\textbf{Ideation and analysis.} The research idea, method design, experimental design, and interpretation of results were conceived by the authors without LLM assistance.
{\small
\bibliographystyle{ieeenat_fullname}
\bibliography{references}
}
\end{document}